\documentclass[runningheads]{llncs}
\usepackage[T1]{fontenc}
\usepackage{amsmath}
\usepackage{graphicx}
\usepackage{booktabs}
\usepackage{multirow}
\usepackage{wrapfig}
\usepackage{hyperref}

\newcommand{\mypar}[1]{\smallskip\noindent\textbf{#1.}}

\begin{document}

\title{Data-Driven Heuristic-Guided Business Process Redesign}

\author{Nan Sai\inst{1}, Karolin Winter\inst{1}\orcidID{0009-0003-8030-2964}, Remco Dijkman\inst{1}\orcidID{0000-0003-4083-0036}}
\institute{Eindhoven University of Technology, Department of Industrial Engineering and Innovation Sciences, Eindhoven, The Netherlands\\
\email{\{n.sai, k.m.winter, r.m.dijkman\}@tue.nl}}
\maketitle
\begin{abstract}
Business process redesign (BPR) has a long-standing catalog of best-practice redesign heuristics, yet, applying them remains a mostly manual task that requires extensive domain knowledge. Despite recent advances using Large Language Models (LLMs) for conversational process model redesign, limitations remain. For example, redesign is user-driven instead of data-driven and is not guided by best practices such as the redesign heuristics. Moreover, only one change can be handled at a time. To address these limitations, given an event log and a BPMN model, we present the DH-BPR framework to realize data-driven heuristic-guided BPR using LLMs. To design DH-BPR, we first conducted an analysis of which redesign heuristics can be automatically detected and applied given information contained in an event log such as the fraction of cases that bypass an activity. Based on this, DH-BPR detects and validates heuristic candidates through a combination of rule-based techniques and an LLM. After that, DH-BPR jointly designs and records an application plan for valid candidates, before finally yielding a redesigned BPMN model and an updated resource model (the way resources execute the process). Evaluation based on simulation using a synthetic benchmark and two real-world event logs shows that the framework's redesign reduces the mean case completion time by up to 61.5\%.

\end{abstract}

\section{Introduction}\label{sec:Introduction}

Business Process Redesign (BPR) constitutes a long-standing, yet crucial, challenge in Business Process Management (BPM)~\cite{Dumas2018}. To facilitate BPR, a plethora of methods have been proposed \cite{DBLP:conf/hicss/GrossMM19}, one of them being the well-known catalog of best-practice redesign heuristics \cite{MANSAR2005457,REIJERS2005283}. However, applying these heuristics in practice remains a mostly manual task for domain experts~\cite{DBLP:journals/dss/FehrerFLRW22}, resulting in substantial effort to detect redesign opportunities and decide which heuristic to apply. Recent advances in Large Language Models (LLMs) raise the prospect of AI-augmented BPM systems where redesign opportunities based on heuristics are automatically detected, planned, and applied~\cite{10.1145/3576047}.

So far, work targeting LLM-based process model redesign focuses on conversational, user-driven scenarios where a chatbot translates a natural-language redesign request into a matching process change pattern, e.g.,~\cite{Klievtsova_2026}. These approaches share, however, three important limitations: i) redesign is user-initiated rather than based on data-driven evidence collected through, e.g., event logs, ii) the LLM is treated as a black-box expert, with limited guidance and checks of its reasoning, and iii) only one change at a time is handled which can lead to overlooking potential conflicts when multiple changes are desirable. 

To address these limitations, we present DH-BPR (Data-driven Heuristic-guided Business Process Redesign). The framework builds upon the well-known redesign heuristics \cite{MANSAR2005457,REIJERS2005283} and takes an event log and a process model expressed in Business Process Model and Notation (BPMN) as input. DH-BPR follows four stages. First, heuristic candidates, i.e., the parts of the process model for which a redesign heuristic could be applied, are detected through rule-based techniques using information extracted from an event log. Second, an LLM validates each candidate based on knowledge on redesign heuristics which we manually derived and prepared from literature as well as it's inherent contextual domain knowledge. Third, an LLM develops a phased redesign plan, grouping the validated heuristic candidates into ordered phases and recording every dropped candidate and why. Finally, the plan is applied to the process model through rule-based transformations. The output of DH-BPR is a redesigned BPMN model and an updated resource model. To illustrate, if the event log shows that most cases bypass an activity, DH-BPR detects that it could be removed, an LLM confirms from the activity's name and context that it is genuinely redundant rather than a mandatory checkpoint, and a rule-based transformation applies the change. Such automatically discovered redesigns can reduce the mean throughput time per case. We evaluate DH-BPR on a synthetic and on two real-world event logs. With the synthetic evaluation we test correctness of DH-BPR's redesign suggestions. For the real-world evaluation we measure reduction in mean throughput time by simulating the original and the redesigned process model.

The paper is organized as follows. Section \ref{sec:Related_Work} elaborates related work, Sec. \ref{sec:Feasibility_Analysis} derives the redesign knowledge DH-BPR builds on. Section~\ref{sec:approach} presents DH-BPR which is evaluated in Sect. \ref{sec:experiments} before the paper concludes in Sect. \ref{sec:conclusion}.

\section{Related Work}\label{sec:Related_Work}
DH-BPR is a data-driven approach which takes, i.a., an event log as input to derive redesign suggestions. One closely related approach is \cite{VANDUN2023113880}. In this approach a generative adversarial network is trained on an event log to generate new process designs. It mainly supports creative idea generation rather than automating heuristic-guided redesign.
Further data-driven approaches fall into rule-based redesign, simulation-based redesign evaluation, pattern-catalog redesign, and assisted redesign.
\cite{VANDERAALST2001451} propose eleven re-engineering rules for \emph{knock-out processes}, that is, processes in which a case must pass a set of checks and is rejected as soon as one check fails. Their rules optimize the ordering, parallelization, and combination of these checks, and are evaluated under the custom Petri-net simulator \emph{ExSpect/KO} whose parameters are supplied by the analyst rather than mined from a log. \cite{Maruster2009} mine an As-Is process model from the event log, simulate an expert-designed To-Be alternative, and compare them across three case studies; we adopt the same As-Is-vs-To-Be paradigm but generate the To-Be model automatically from the heuristic catalog rather than relying on expert design. \cite{10.1007/978-3-642-21759-3_7} pair a catalog of optimization patterns with audit-trail-based data analysis to assist analysts in selecting and applying the most fitting pattern for a given situation. More recently, \cite{DBLP:journals/dss/FehrerFLRW22} conceptualize \emph{assisted} BPR as four levels of automation supported by a reference architecture, with domain expert input still essential at all four levels; in DH-BPR such input is optional rather than required.
None of these approaches uses LLMs or models redesign conflicts. 

Recent research explored how to use LLMs for BPR. For example, \cite{Klievtsova_2026} present conversational process model redesign. The LLM identifies which change pattern from the catalog of \cite{10.1007/978-3-540-72988-4_40} matches the request, derives its parameters and applies it. This work is user-driven, not data-driven, and handles one change at a time.

Lastly, from a methodology point of view, DH-BPR aligns with recent work that grounds LLM reasoning in explicit knowledge structures such as knowledge graphs~\cite{edge2025localglobalgraphrag,10387715}. To guide the final planning step, we developed a graph based on literature analysis reflecting relations and conflicts between redesign heuristics.

\section{Deriving the Redesign Knowledge for DH-BPR}\label{sec:Feasibility_Analysis}
Before presenting the DH-BPR framework (Section~\ref{sec:approach}), we describe how we developed the redesign knowledge it relies on. Starting from the redesign heuristics~\cite{MANSAR2005457,REIJERS2005283}, our aim is to make them applicable from an event log and a BPMN model. We first conducted an automatability analysis selecting the heuristics an event log and a BPMN model can support (Sect. \ref{subsec:automatability}). For each selected heuristic, we then derive the knowledge DH-BPR requires: detection rules (Sect. \ref{subsec:detection-rules}), an evidence interpretation guide for validation (Sect. \ref{subsec:evidence-glossary}), and a Heuristic Interaction Graph for redesign planning (Sect. \ref{subsec:hi-graph}). All detailed material behind these steps is available in a public repository.\footnote{\url{https://github.com/Christoph9007/bpr-llm}\label{fn:repo}}

\subsection{Automatability Analysis}\label{subsec:automatability}
For each of the $29$ heuristics, \cite{MANSAR2005457,REIJERS2005283} details the situation in which a heuristic applies (applicable situation) as well as the transformation it prescribes, i.e., the adaption(s) performed on the BPMN model or the resources executing the process. For each heuristic, we analyzed whether its detection and application can be automated given an event log and a BPMN model. We distinguish direct attributes, namely activity name, case identifier, timestamp, resource, and lifecycle transition, and derived attributes, computed from the direct ones through standard process analysis techniques. The derived attributes include, for example, activity frequency, waiting time, handover count, and dependency strength.

We label a heuristic as \emph{automatically detectable} when its applicable situation can be matched against either direct or derived attributes. For example, Activity Elimination applies to a redundant activity, one that is bypassed by most cases and waits far longer than it runs. Both signals, its bypass ratio and its waiting-to-duration ratio, are quantifiable from the log, so it is automatically detectable.

We label a heuristic as \emph{automatically applicable} when its prescribed transformation reduces to a fixed operation on the BPMN model or on the resources executing it, with the activities the operation targets determined from the same direct and derived attributes. Applying Activity Elimination removes the activity that detection already flagged, so no further input is needed to locate it. Outsourcing, by contrast, requires external context that neither the log nor the BPMN model contain, so it is not detectable and not applicable.

This analysis revealed that 16 of the 29 heuristics are both automatically detectable and applicable. The full classification of all 29 heuristics is documented in the repository.\textsuperscript{\ref{fn:repo}} From those 16 we select \textsl{Activity Elimination}, \textsl{Resequencing}, \textsl{Case Assignment}, and \textsl{Parallelism} as those are among the top-ten most-used heuristics derived empirically by~\cite{MANSAR2005457}. Furthermore, we select \textsl{Knock-out}, \textsl{Exception}, and \textsl{Case-based Work} to obtain a representative set of seven heuristics spanning both classes of automatable transformation, namely changes to the BPMN model and changes to the resources executing it.

\subsection{Heuristic Candidate Detection Rules}\label{subsec:detection-rules}
All direct and derived attributes are organized into four \emph{evidence families}: control-flow, temporal, resource, and batch. The first three are standard process perspectives. Batch was added because case-based work hinges on whether an activity's work arrives and is processed in batches, a distinct signal the other three cannot derive and one whose redesign can yield the largest gains, since a batch wait is avoidable idle time. The features within these families come from the redesign literature~\cite{MANSAR2005457,REIJERS2005283}, each condition of a heuristic's applicable situation mapped to one or more log features.

For each of the seven selected heuristics we manually derive a \emph{per-heuristic rule} over the direct and derived features computed from the event log. We illustrate this using Activity Elimination, which targets redundant or non-value-adding activities. For each activity $a$, the rule reads three features: i) the bypass ratio $b(a)$, i.e., the fraction of cases that bypass $a$, ii) the mean execution duration $d(a)$, and iii) the ratio $r(a) = \frac{w(a)}{d(a)}$ of $a$'s mean waiting time to its duration.

A raw feature value says little until benchmarked against the other activities. A rule therefore turns each feature into an \emph{evidence condition}, a threshold the value must pass, and, wherever all of its conditions hold, produces one \emph{heuristic candidate}, a location where the heuristic could apply together with the evidence supporting it. A rule may also carry a \emph{structural condition} read from the BPMN model, for example that two activities are in a directly-follows relation.

The thresholds are mostly percentile-based against the log, so a feature is judged relative to the other activities rather than against a fixed value. Where percentile ranking is not feasible, absolute values are used. The rules are deliberately permissive, to avoid missing suitable candidates. Conditions and thresholds for all seven heuristics are documented in the project repository. \textsuperscript{\ref{fn:repo}}

Returning to Activity Elimination, its rule fires when all three conditions hold: $b(a) > 0.5$ (more cases bypass $a$ than traverse it), $r(a)$ above the activity-wise median of $66$ (the wait is large relative to the work), and $d(a)$ below the 75th percentile of durations, $0.27$ h (the activity itself is not slow and substantial). The duration condition is a guard, waived when $b(a)$ is very high, since a heavily bypassed activity is a strong elimination candidate regardless of its execution time. For an example activity with $b = 0.998$, $r = 92$, and $d = 0.20$ h, all three conditions pass and the activity is flagged as an Activity Elimination candidate.

\subsection{Evidence Interpretation Guide}\label{subsec:evidence-glossary}
The percentile thresholds of Section~\ref{subsec:detection-rules} are relative. They compare an activity's feature value against the other activities in the same log, so a candidate can be detected, yet still be weak in absolute terms. The evidence interpretation guide adds the absolute reference that such a relative threshold lacks. It is a manually derived table, one per heuristic, built from the redesign literature~\cite{MANSAR2005457,REIJERS2005283}. An \emph{evidence field} is one numeric field of a candidate's evidence, such as \texttt{bypass\_ratio} for Activity Elimination. For each field, the guide gives two aspects. First, the field's meaning, namely, what the number measures and whether a higher or lower value favors the heuristic. For \texttt{bypass\_ratio}, the fraction of cases that skip the activity, a higher value points toward elimination, since it indicates this activity could be structurally optional. The second aspect is an absolute acceptance band that fixes how high is high enough by grading the value as supporting, weak, or insufficient. For \texttt{bypass\_ratio}, supporting sits above $0.9$, so an activity with $b = 0.95$ falls in the supporting band. The guide thus makes the rule-based detection interpretable in absolute terms, complementing it rather than restating it.

\subsection{Heuristic Interaction Graph}\label{subsec:hi-graph}
The guide supports assessing each candidate in isolation. But two candidates that are each valid can still conflict. For example, an Activity Elimination and a Knock-out that target the same activity are each justified on their own, yet eliminating the activity removes what the Knock-out reordering needs. Such candidates must therefore be weighed against each other. The Heuristic Interaction Graph (HI-Graph) captures these interactions. It is an undirected graph on the seven heuristics, manually derived from literature~\cite{MANSAR2005457,REIJERS2005283}. It attaches a \emph{type relation} to each of the $\binom{7}{2} = 21$ pairs of the seven heuristics. \texttt{HARD\_CONFLICT} marks a structurally incompatible pair, as above, where applying both would break the model since one removes or moves an activity the other needs; at most one can be kept. \texttt{WEAK\_CONFLICT} marks a compatible pair whose effects still interact, so both can be kept only with coordination. \texttt{NEUTRAL} marks an independent pair. Of the 21 heuristic pairs, 5 are \texttt{HARD\_CONFLICT}, 13 are \texttt{WEAK\_CONFLICT}, and 3 are \texttt{NEUTRAL}.

\section{The DH-BPR Framework}\label{sec:approach}

\begin{figure}[!htbp]
\centering
\includegraphics[width=0.76\textwidth]{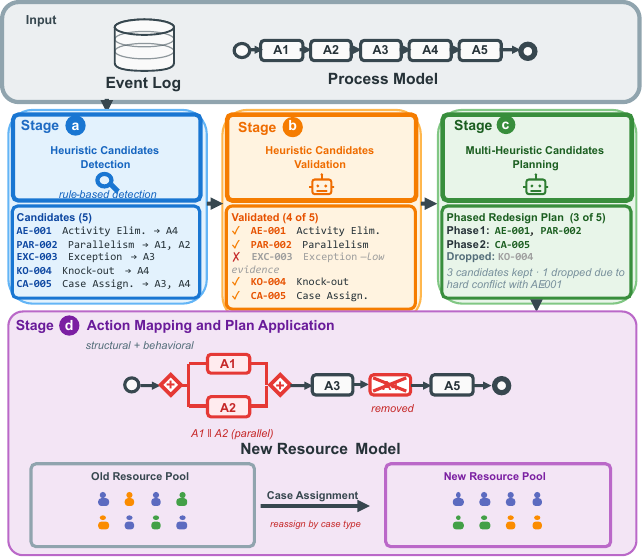}
\caption{Overview of DH-BPR}
\label{fig:framework-overview}
\end{figure}

Figure \ref{fig:framework-overview} outlines our data-driven heuristic-guided automated BPR framework DH-BPR. It takes an event log and a BPMN model as input, and outputs a redesigned BPMN model together with an updated resource model, the way resources execute the process. The resource model comprises the resource pool (which resources can perform which activities) and the work-execution rules (for example, case by case or in batches). DH-BPR has four stages (a)--(d). Stage~(a) uses a rule-based detector to identify heuristic candidates (Section~\ref{subsec:detection-rules}). Stage~(b) uses an LLM to validate these candidates, keeping the accepted ones. Stage~(c) resolves conflicts between accepted candidates and orders them into a redesign plan. Stage~(d) maps each candidate to a concrete edit operation and applies the edits to the BPMN model and the resource model. Stages (a)-(d) are detailed below, illustrated with the example in Fig. \ref{fig:framework-overview}.

\subsection{Stage (a): Heuristic Candidate Detection}\label{sec:stage-a}
This stage tests every location against the per-heuristic rules of Section~\ref{subsec:detection-rules}, using the features computed from the event log and the BPMN model, and outputs a heuristic candidate wherever a rule fires. On the running example, it outputs five such candidates (Fig.~\ref{fig:framework-overview}), which Stage~(b) validates next.

Besides its heuristic, target activity, evidence, and local process fragment, each candidate also carries an \emph{additional context} note that the detector derives from the same event log. The note paraphrases the candidate's numbers in words, for example that the activity is reworked in $12\%$ of cases, and appends a fixed reminder that external constraints such as compliance rules are not in the log. The additional context is therefore derived information, not a third input.

\subsection{Stage (b): Heuristic Candidates Validation}\label{sec:phase1-validation}

\begin{figure}[!htbp]
\centering
\includegraphics[width=0.85\textwidth]{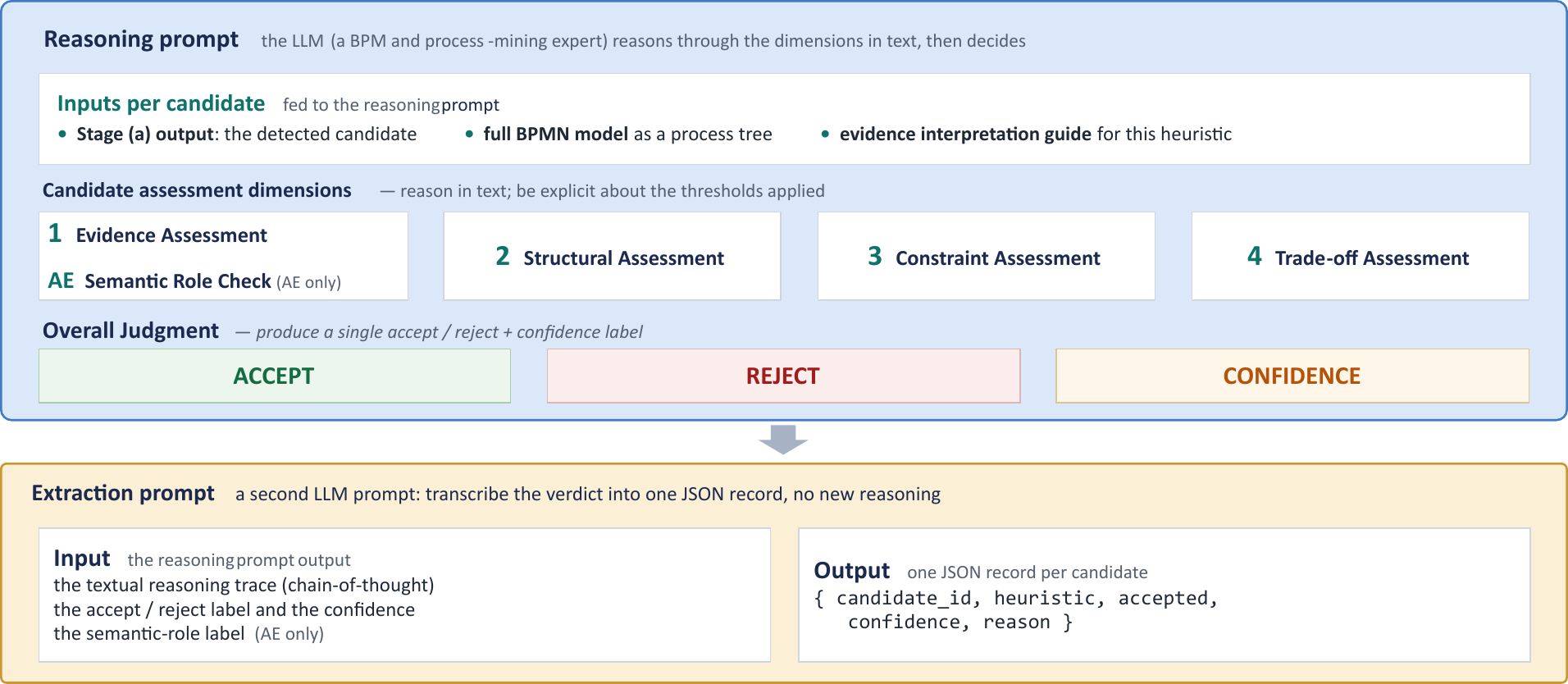}
\caption{Validation prompts: reasoning prompt (top) and extraction prompt (bottom)}
\label{fig:phase1-prompt}
\end{figure}

Determining candidates from numeric evidence is a useful starting point, but a candidate's validity can also depend on information beyond the numbers, such as the meaning of an activity in its business context, its position in the process model, and its relations to other activities. Adding such context as extra input would complicate DH-BPR's inputs and their processing. As LLMs have shown their potential for semantic business process management tasks \cite{DBLP:journals/procsci/RebmannSGA25}, we decided to integrate them into DH-BPR to address the above challenges. The validation step is illustrated in Fig.~\ref{fig:phase1-prompt} and consists of two LLM prompts, one for reasoning and one for extracting a JSON record. Each prompt is invoked separately per heuristic candidate. The output consists of a refined and validated list of candidates, each labeled as \emph{accepted} or \emph{rejected}.

\mypar{Reasoning Prompt} We employ chain-of-thought (CoT) prompting, i.e., we ask the LLM to explicate its reasoning and then give a verdict, rather than reply with only accept or reject. This well-established technique has been shown to substantially improve performance on reasoning tasks~\cite{10.5555/3600270.3602070}.  

The reasoning prompt receives the following inputs. First, the candidate itself as obtained from Stage~(a) containing the heuristic, its target activity, the candidate's evidence, its process fragment (a local BPMN subgraph, possibly empty), and its additional context. Second, the full BPMN model, rendered as a process tree, i.e., a block-structured tree representation that serializes to text and so embeds in the prompt more cleanly than a BPMN diagram.  

Third, the per-heuristic evidence interpretation guide (Section~\ref{subsec:evidence-glossary}), which gives each evidence field a meaning and an absolute acceptance band. We hand the guide to the LLM inside the prompt rather than use it as a hard pre-filter. Evidence Assessment is only one of the candidate assessment dimensions introduced next, and the verdict weighs them together, so a threshold filter would drop a candidate on a weak number before the LLM could weigh the other dimensions. Inside the prompt, the LLM's CoT output also shows each value, the band it falls in, and the verdict it supports.

The next part of the reasoning prompt gives the LLM four candidate assessment dimensions, boxes 1 to 4 in Fig.~\ref{fig:phase1-prompt}, each catching a different way a candidate can pass the detector yet still be a poor target. Evidence Assessment (box 1) asks whether the numbers really support the change. It checks the candidate's evidence against the guide's bands. Structural Assessment (box 2) asks whether the local BPMN allows the change. It checks the candidate's process fragment against the full process tree. Constraint Assessment (box 3) asks whether outside rules permit the change. The event log holds no compliance or regulatory information, so the LLM judges from the activity's meaning and its own knowledge. This mostly raises soft concerns that lower confidence and flag the candidate for human review. Only rarely does it produce a hard block that forces rejection, a limit we revisit in Section~\ref{sec:conclusion}. Trade-off Assessment (box 4) asks whether the gain is worth the cost, weighing its impact on the four Devil's Quadrangle dimensions. For Activity Elimination only, the prompt adds a sub-step, Semantic Role Check (labeled AE inside box 1), that labels the target activity redundant, conditional quality control, or mandatory. It is needed because a high bypass ratio is ambiguous, since an equally-skipped activity could be a mandatory compliance check. In the running example it labels A4 redundant, so AE-001 is accepted.

Finally, the four answers feed an Overall Judgment, the accept-or-reject label plus a confidence of high, medium, or low. Acceptance needs all four dimensions to pass and a non-trivial improvement; rejection follows if any dimension fails, the magnitude is below threshold, or a hard constraint blocks.

\mypar{Extraction Prompt}
The reasoning prompt's output is in textual form. A follow-up extraction prompt therefore has the LLM rewrite that free text as JSON, so the results can be used in the subsequent stages of DH-BPR.

\mypar{Running Example} Figure \ref{fig:framework-overview} illustrates the validation step's judgments on the five candidates of the running example. PAR-002 proposes running A1 and A2 in parallel; the four dimensions all pass and the validation step accepts at high confidence. EXC-003 proposes isolating A3 as a rare exception path, but A3's rare-path frequency falls in the insufficient band of the guide, so the validation step rejects it on Evidence Assessment. The remaining three candidates (AE-001, KO-004, CA-005) clear all four dimensions and pass to Stage~(c) for planning.

\subsection{Stage (c): Multi-Heuristic Candidates Planning}\label{sec:phase2-planning}
Planning takes the candidates that Stage~(b) accepted and turns them into a \emph{Phased Redesign Plan}. It keeps only a subset of the accepted candidates, splits that subset into ordered phases applied one after another, and records for every dropped candidate why it was excluded. It must settle two things. First, candidates can conflict, since one accepted in isolation may, for example, remove an activity that a later Knock-out reordering depends on. Second, the user supplies a goal, balanced by default, for example optimizing the time dimension of the Devil's Quadrangle. A deterministic step pre-computes both for the candidate set. It determines which candidates conflict based on the HI-Graph (Section~\ref{subsec:hi-graph}), and scores each candidate against the goal via a per-heuristic impact vector below. An LLM then turns the candidates and these inputs into the plan.

\mypar{Conflicts} For a candidate set, a deterministic step looks up each candidate pair in the HI-Graph's type relations (Section~\ref{subsec:hi-graph}) and outputs the \emph{interaction report}, the list of pairs that actually conflict, namely the \texttt{HARD\_CONFLICT} and \texttt{WEAK\_CONFLICT} pairs that fire on this set. \texttt{NEUTRAL} and non-firing pairs are left out. The LLM later arbitrates this list instead of rediscovering the conflicts itself.

\mypar{Goal Scores} Separately from the HI-Graph, DH-BPR attaches an \emph{impact vector} to each heuristic, its effect on the four Devil's Quadrangle dimensions, time, cost, quality, and flexibility. Each entry takes one of five ordinal levels, from strong positive to strong negative ($+2$ to $-2$). The signs follow Table~1 of~\cite{REIJERS2005283}. We set a level to strong when the transformation directly removes, adds, or replaces the work that drives a dimension, and to weak when it only reshuffles existing work. Activity Elimination, for example, scores strong positive on time and cost and weak negative on flexibility. The user supplies a \emph{goal} at run time, a \emph{weight vector} that says how much each of the four Devil's Quadrangle dimensions matters. A deterministic step then gives each candidate a \emph{goal score}, the dot product of the normalized goal weights with the heuristic's impact vector.

\mypar{The LLM Planner}\label{subsec:goal-driven-planning} The planner is an LLM that uses a similar two-prompt structure as the validation step: a reasoning prompt and an extraction prompt. What changes is the scope. Validation judged one candidate at a time, whereas the planner sees the whole accepted set of candidates at once. The reasoning prompt receives, as inputs, the process tree, the accepted candidate set with its evidence, the interaction report, and the goal scores. It then casts the LLM as a planner and walks it through four planning dimensions.

The four planning dimensions have different aims. Conflict Arbitration resolves each entry in the interaction report. For a \texttt{HARD\_CONFLICT} it keeps at most one of the two candidates, justified by goal scores and evidence; a \texttt{WEAK\_CONFLICT} keeps both unless the evidence demands dropping one. Phase Assignment groups the retained candidates into ordered phases. The only hard rule is that a retained Exception candidate goes in the earliest phase. Goal Alignment checks the plan against the goal, weighing negative goal scores against retention. Overall Coherence runs a final review, confirming that no \texttt{HARD\_CONFLICT} pair is both kept, every \texttt{WEAK\_CONFLICT} is addressed, and the phase order is consistent. The extraction prompt then records the result, the Phased Redesign Plan, as JSON.

\mypar{Running Example}\label{subsec:phased-plan-example} Stage~(c) of Fig.~\ref{fig:framework-overview} walks through the four accepted candidates of the running A1--A5 example (AE-001, PAR-002, KO-004, CA-005) under the goal \emph{optimize-time}. The interaction report lists two conflicts, a \texttt{HARD\_CONFLICT} between AE-001 and KO-004 (both target A4, so removing it via Activity Elimination invalidates KO-004's reordering) and a \texttt{WEAK\_CONFLICT} between AE-001 and CA-005 (the eliminated activity is also pooled by CA-005). The LLM arbitrates the HARD entry by keeping AE-001, whose goal score under \emph{optimize-time} is~$+2$ against KO-004's~$+1$; the remaining three candidates clear Goal Alignment, and Phase Assignment places AE-001 and PAR-002 in Phase~1 and CA-005 in Phase~2. The Phased Redesign Plan is the input to Stage~(d)'s action mapping.

\subsection{Stage (d): Action Mapping and Plan Application}\label{sec:stage-c}
The final stage of DH-BPR, Stage~(d) in Fig.~\ref{fig:framework-overview}, consists of action mapping and plan application. Both are deterministic. Each kept candidate of the Phased Redesign Plan is first turned into a concrete \emph{action}, one per edit operation. Each edit operation realizes the transformation that one of the seven heuristics prescribes and corresponds to an established process-model change pattern~\cite{10.1007/978-3-540-72988-4_40}. We implement edit operations for structural process changes (removing an activity, inserting an activity at a given position, wrapping a sequential pair in parallel gateways, and wrapping a branch into a sub-process) as well as behavioral ones. Structural edits alter the BPMN process model while behavioral ones edit the resource model (either its resource pool or its work-execution rules). For example, AE-001 (eliminate A4) maps to a structural action removing that activity from the BPMN model. 
The action plan is then applied one action at a time, in the phase order that planning produced. Each edit mutates the model in place, so later edits act on the result of earlier ones. This is why planning fixes the order, for example applying an Exception isolation before edits that would otherwise target the branch it splits off. A structural edit can also leave the resource model inconsistent with the edited model, since a removed activity still has resources assigned to it. DH-BPR therefore updates the resource model after each structural edit, so that the structural and behavioral outputs stay consistent.

\section{Evaluation}\label{sec:experiments}
DH-BPR was prototypically implemented and evaluated on synthetic and real-world datasets. The source code and datasets are publicly available.\textsuperscript{\ref{fn:repo}} All LLM components use \texttt{gpt-5.5} at temperature $T = 1.0$. The synthetic dataset consists of 17 scenarios with manually derived ground-truth labels for the detection, validation, and planning stages of DH-BPR (see Sect. \ref{subsec:eval-synthetic}). It serves for evaluating the quality of DH-BPR's outputs in a controlled setting. The two real-world datasets are based on the BPI Challenge 2012~\cite{https://doi.org/10.4121/uuid:3926db30-f712-4394-aebc-75976070e91f} and BPI Challenge 2017~\cite{https://doi.org/10.4121/uuid:5f3067df-f10b-45da-b98b-86ae4c7a310b} event logs. These serve to test whether DH-BPR's redesign suggestions translate into measurable process improvements in real-world scenarios (see Sect. \ref{subsec:eval-real}).

\subsection{Synthetic Benchmark}\label{subsec:eval-synthetic}
Real-world event logs carry no ground-truth annotation of which heuristic applies where, so they cannot reliably evaluate DH-BPR's detector, validation, and planning stages. We therefore built a synthetic benchmark of 17 scenarios in which the seven selected heuristics are applicable by construction.

\mypar{Dataset Construction} We build each scenario by hand. A scenario has three parts: a BPMN process model with single-letter activity labels ``A'' to ``H'', an event log of 200 traces, and expected ground-truth labels. To generate the log, we write a small set of valid trace variants through the model and sample the 200 cases from them by frequency weight. We set per-activity waiting time, duration, and case inter-arrival during sampling, so the target heuristic's signal is present by construction and the detector's percentile thresholds are clearly exceeded. The ground-truth labels record, for every (heuristic, target) pair the scenario covers, three Boolean fields: whether the detector should emit the candidate, whether validation should accept it, and whether planning should retain it. The benchmark checks only whether DH-BPR's decisions match these labels, not the redesigned model itself, which we assess on the real-world logs (Section~\ref{subsec:eval-real}).

\mypar{Scenario Categories} The 17 scenarios fall into three categories, each with a distinct purpose. The first two isolate a single heuristic; the third combines several. The 8 \textsl{single-positive scenarios} make exactly one heuristic applicable at one target activity and test whether the detector and validation flag that true positive. There are 8 because Activity Elimination is tested on both a short and a long process, the other 6 once each. The 6 \textsl{all-negative scenarios} contain no true positive and so measure the specificity of the detector and validation. One is a pure-negative control, a clean process on which no heuristic should fire, testing specificity for all seven at once. The other five are near-misses, each a process that looks like one heuristic applies (Activity Elimination, Resequencing, Knock-out, Parallelism, or Case-based Work) but breaks one critical condition of its rule, testing whether the detector resists a tempting false positive that the control cannot. Exception and Case Assignment have no dedicated near-miss; their false positives instead surface across the other scenarios. The 3 \textsl{multi-positive scenarios} place several true positives in one scenario, some of them in conflict, and test whether planning handles the candidates jointly.

\mypar{Evaluation Metrics} Each (heuristic, target) pair yields, for each of the three stages, one outcome (TP, FP, FN, or TN) against the ground-truth label, from which we compute precision, recall, and F1-score. The 17 scenarios contribute $308$ such pairs, scored identically at every stage, so a category's three rows are directly comparable. A candidate whose set-valued target matches no labelled pair, as for Case Assignment and Knock-out, counts as a false positive.

\mypar{Evaluation Results} Table \ref{tab:synthetic-categories} reports precision, recall, and F1 per scenario category and, in the bottom row, aggregated over all 17 scenarios. We evaluate the detection, validation, and planning stages. Plan application makes no accept-or-reject decision, so we evaluate it instead on the real-world logs (Section~\ref{subsec:eval-real}), where the model it produces is run in simulation against the baseline. Overall, the F1-score rises from $0.50$ at detection to $0.63$ at validation, which removes 10 of the 22 false positives while keeping every true positive; planning ends at $0.61$, trading two true positives for one false positive.

In the 8 single-positive scenarios, recall remains at $1.00$ through detection and validation, so all 8 genuine opportunities are found by the detector and kept by the validator stage. Precision increases from $0.47$ to $0.62$ between these stages, because validation rejects four of the nine spurious candidates that the permissive detector emitted alongside the true ones. At planning, precision reaches $0.64$, but recall drops to $0.88$.
This can be explained by a spurious Case Assignment candidate outscoring a correct Parallelism in a \texttt{HARD\_CONFLICT}, so the planner keeps the wrong candidate and drops the genuine one. In the 6 all-negative scenarios every emission is by construction a false positive: the detector raises 8 on these plausible-looking negatives, of which the later stages remove 3. The 3 multi-positive scenarios are the most challenging, and their recall stays at $0.67$ at detection and validation. This ceiling is set at detection and cannot be lifted later. When two heuristics target the same activity, their overlapping evidence conditions let only one rule fire, so the second genuine candidate is never emitted. Because validation and planning can only remove candidates, never add them, this missed true positive is unrecoverable, which caps recall at every stage. Beyond the per-category figures, the residual error concentrates on two heuristics, Exception and Case Assignment, which together account for 11 of the 12 false positives that survive validation. The over-firing is structural. At detection the rule fires on any rare, slower variant branching off the dominant path, with no test to tell a genuine exception from a benign minority path, so validation receives candidates it cannot separate on evidence alone. That validation accepts them here is partly a benchmark artifact: single-letter labels give the LLM no semantic cue, and distractor paths carry throughput gaps large enough to look like real exceptions. On the real-world logs, where activities have meaningful names, validation does reject Exception candidates (all four raised on BPI Challenge 2017), so the failure is far less visible there, though real-world Exception precision stays unquantified for lack of ground truth. Case Assignment over-fires too, proposing a resource-pool consolidation in 8 scenarios where none holds; validation accepts 6, lacking any organisational constraint against one.

\begin{table}[!htbp]
\centering
\caption{Results by category and stage; counts are TP/FP/FN/TN. Of the $308$ scored pairs only 14 are positive; precision, recall, and F1 are undefined ($-$) without one.}
\label{tab:synthetic-categories}
\scriptsize
\renewcommand{\arraystretch}{0.8}
\setlength{\tabcolsep}{5pt}
\begin{tabular}{@{}llcccc@{}}
\toprule
Category & Stage & TP/FP/FN/TN & Precision & Recall & F1 \\
\midrule
\multirow{3}{*}{Single-positive ($8$)}
  & Detection  & $8/9/0/139$ & $0.47$ & $1.00$ & $0.64$ \\
  & Validation & $8/5/0/143$ & $0.62$ & $1.00$ & $0.76$ \\
  & Planning   & $7/4/1/144$ & $0.64$ & $0.88$ & $0.74$ \\
\midrule
\multirow{3}{*}{All-negative ($6$)}
  & Detection  & $0/8/0/93$ & $-$ & $-$ & $-$ \\
  & Validation & $0/5/0/96$ & $-$ & $-$ & $-$ \\
  & Planning   & $0/5/0/96$ & $-$ & $-$ & $-$ \\
\midrule
\multirow{3}{*}{Multi-positive ($3$)}
  & Detection  & $4/5/2/40$ & $0.44$ & $0.67$ & $0.53$ \\
  & Validation & $4/2/2/43$ & $0.67$ & $0.67$ & $0.67$ \\
  & Planning   & $3/2/1/45$ & $0.60$ & $0.75$ & $0.67$ \\
\midrule
\multirow{3}{*}{Overall ($17$)}
  & Detection  & $12/22/2/272$ & $0.35$ & $0.86$ & $0.50$ \\
  & Validation & $12/12/2/282$ & $0.50$ & $0.86$ & $0.63$ \\
  & Planning   & $10/11/2/285$ & $0.48$ & $0.83$ & $0.61$ \\
\bottomrule
\end{tabular}
\end{table}

\subsection{Real-World Event Logs}\label{subsec:eval-real}
We now evaluate whether DH-BPR's redesigns improve real-world processes, using the BPI Challenge 2012~\cite{https://doi.org/10.4121/uuid:3926db30-f712-4394-aebc-75976070e91f} and 2017~\cite{https://doi.org/10.4121/uuid:5f3067df-f10b-45da-b98b-86ae4c7a310b} event logs. To measure improvement, we simulate the baseline and redesigned process model and compute Key Performance Indicators (KPIs). The KPIs follow the process-mining-based framework \cite{CHO201792} and cover the time and cost dimensions of the Devil's Quadrangle. Quality and flexibility are out of scope, as simulation cannot capture them.

\mypar{Dataset Construction} We first mine simulation models from both logs with Simod~\cite{CHAPELACAMPA2025102157}, which discovers a simulation model from an event log. Its output is our baseline, the BPMN process model with the resource model and the timing and branching parameters needed to simulate it. We run the model in Prosimos~\cite{10.1007/978-3-031-26886-1_23}, a simulation engine that replays it to generate an event log from which we compute the KPIs. To account for the stochasticity of simulation, we simulate each model $20$ times and average the resulting KPIs. Neither simulation includes calendar-level delays, because the log's between-activity gaps are heavy-tailed (mostly short, with rare multi-day pauses) and Simod's delay model cannot fit this shape. Since both share the same Simod-mined model, their comparison is unaffected. Case-based Work's redesign eliminates a batch wait the Simod baseline omits, so for each retained candidate we insert a BPMN timer event before the target activity, with duration equal to its mean batch interval. A timer pauses the case without occupying a resource, matching a batch wait.

\mypar{Evaluation Metrics and KPIs} Table~\ref{tab:realworld-kpi} reports relevant KPIs and their differences $\Delta$ between baseline and redesign. \emph{Throughput} is the per-case span from a case's first activity enable time to its last activity end time. We summarize its distribution across cases by the mean, the median, and the 99th percentile. \emph{Waiting time per case} sums each event's waiting time within a case, then averages across cases. \emph{Waiting time per event} averages those event waiting times across events. \emph{Resource utilization} is the mean fraction of the simulated horizon for which each resource is busy. Throughput and the two waiting measures capture the \emph{time} dimension of the Devil's Quadrangle, while resource utilization is our proxy for the \emph{cost} dimension, since resource usage is the main cost driver when no monetary cost rates are available in the log. All changes reported below except the BPI 2017 resource utilization are significant at $p < 0.01$ (Welch two-sample $t$-test over the 20 runs per configuration).

\begin{table}[!htbp]
\centering
\caption{KPI means of 20 simulation runs for BPIC 2012 and 2017; standard deviation in parentheses; times in hours; $\Delta$ is the relative change between baseline and redesign, computed from unrounded means. $^{\dagger}$ marks the change not distinguishable from zero.}
\label{tab:realworld-kpi}
\scriptsize
\renewcommand{\arraystretch}{0.8}
\setlength{\tabcolsep}{4pt}
\begin{tabular}{@{}llrrr@{}}
\toprule
Dataset & KPI & Base & Redesigned & $\Delta$ (\%) \\
\midrule
\multirow{6}{*}{\emph{BPI 2012}}
  & Throughput mean / case     & $5.475\,(0.165)$     & $4.998\,(0.183)$     & $-8.7$ \\
  & Throughput median / case   & $2.006\,(0.120)$     & $1.697\,(0.162)$     & $-15.4$ \\
  & Throughput p99 / case      & $37.87\,(1.02)$      & $32.02\,(2.41)$      & $-15.4$ \\
  & Avg waiting time per case  & $4.206\,(0.166)$     & $4.051\,(0.190)$     & $-3.7$ \\
  & Avg waiting time per event & $0.2136\,(0.0089)$   & $0.2762\,(0.0134)$   & $+29.3$ \\
  & Avg resource utilization       & $0.14504\,(0.00920)$ & $0.08281\,(0.00433)$ & $-42.9$ \\
\midrule
\multirow{6}{*}{\emph{BPI 2017}}
  & Throughput mean / case     & $23.983\,(0.434)$    & $9.225\,(0.440)$     & $-61.5$ \\
  & Throughput median / case   & $14.437\,(0.306)$    & $3.460\,(0.307)$     & $-76.0$ \\
  & Throughput p99 / case      & $130.11\,(5.31)$     & $65.72\,(8.64)$      & $-49.5$ \\
  & Avg waiting time per case  & $10.216\,(0.250)$    & $5.107\,(0.301)$     & $-50.0$ \\
  & Avg waiting time per event & $0.16891\,(0.00471)$ & $0.08420\,(0.00525)$ & $-50.1$ \\
  & Avg resource utilization       & $0.1663\,(0.0293)$   & $0.1589\,(0.0230)$   & $-4.4^{\dagger}$ \\
\bottomrule
\end{tabular}
\end{table}

\mypar{BPI Challenge 2012} Table \ref{tab:realworld-kpi} shows a decrease in throughput: mean by $8.7\%$, and median and 99th percentile both by $15.4\%$. Waiting time per case drops only slightly ($-3.7\%$), waiting time per event even rises ($+29.3\%$), and resource utilization decreases by $42.9\%$. These numbers are driven by the four candidates that planning retained: one Activity Elimination that removes a phone-call-back step on offers, two Resequencing candidates that reposition the application-completion step and a phone-call-back step on incomplete files, and one Exception that isolates a rare-path branch into a sub-process. Two of these entries look like deterioration but are side effects of how the redesign reshapes the waiting distribution. The increasing per-event waiting time and the steep utilization drop both come from the Activity Elimination, which removes an activity that runs about $3.5$ times per case, each visit with a low waiting time. The deleted low-wait events had been dragging the per-event mean down, so removing them lifts the average of what remains, while each case still sheds total waiting time and the freed resource lowers utilization. The same redistribution surfaces at the pre-acceptance step. Its per-event waiting time increases by $35\%$, yet its median waiting time stays at zero. Most cases still pass through instantly, but upstream speed-ups push a few past closing time, so they wait until the next workday. 

\mypar{BPI Challenge 2017} Table \ref{tab:realworld-kpi} shows substantial reductions: throughput drops by $61.5\%$ in the mean, $76.0\%$ in the median, and $49.5\%$ at the 99th percentile. Waiting time per case and per event both almost halve ($-50.0\%$ and $-50.1\%$), while resource utilization shows no distinguishable change ($-4.4\%$). These results can be explained by the four retained candidates: two Case-based Work candidates on the batched application-validation and application-completion steps, one Activity Elimination that removes a lead-handling step, and one Case Assignment that consolidates the resource pool of four downstream application and offer activities. Most of the reduction is due to the two Case-based Work candidates, whose mean batch intervals are $1.03$~h and $0.81$~h; each case visits them multiple times, so removing the batch waits saves several hours per case. The Activity Elimination and Case Assignment candidates contribute less, the latter mainly by reducing service time at the follow-up call step. That resource utilization shows no distinguishable change ($95\%$ CI $[-14.3\%, +5.5\%]$) is itself informative. The resources stay about as busy as in the baseline, so the speed-up comes from cutting waiting time, not from idle resources.

\section{Conclusion}\label{sec:conclusion}
This paper presented DH-BPR, a framework for data-driven heuristic-guided business process redesign. The framework builds upon knowledge on redesign heuristics which was manually derived and translated into computational evidence. DH-BPR uses this evidence to suggest a redesigned BPMN and resource model based on, i.a., event log information. In particular, seven rule-based detectors propose heuristic candidates which are checked by an LLM-based validation step. The planning step jointly arbitrates validated candidates under HI-Graph rules, and the final plan application step emits a redesigned BPMN model and resource model. Evaluation on 17 synthetic scenarios confirmed the quality of results produced by DH-BPR. In addition, an evaluation featuring process simulation on two real-world event logs showed that DH-BPR's redesign lowers the mean throughput time per case on both real-world logs ($-8.7\%$ on BPI Challenge 2012, $-61.5\%$ on BPI Challenge 2017), with the Exception heuristic surfacing as DH-BPR's structural weak spot.
Right now, DH-BPR only considers an event log and BPMN model as inputs. However, as our automatability analysis demonstrated, regulatory or policy information is valuable. The validation step of DH-BPR can only approximate such contextual knowledge through the LLM's prior, not validated compliance knowledge. DH-BPR's candidates therefore require domain expert review before deployment. The KPI comparison also relies on simulation, whose validity is bounded by the fidelity of the mined baseline. Complementary validation through deployment studies, expert assessment, or comparison against expert-designed To-Be models is left to future work. We also plan improvements for the Exception heuristic and to extend DH-BPR to more of the automatable heuristics.

\bibliographystyle{splncs04}
\bibliography{references}

\begin{thebibliography}{10}
\providecommand{\url}[1]{\texttt{#1}}
\providecommand{\urlprefix}{URL }
\providecommand{\doi}[1]{https://doi.org/#1}

\bibitem{CHAPELACAMPA2025102157}
Chapela-Campa, D., López-Pintado, O., Suvorau, I., Dumas, M.: Simod: Automated
  discovery of business process simulation models. SoftwareX  \textbf{30},
  102157 (2025). \doi{10.1016/j.softx.2025.102157}

\bibitem{CHO201792}
Cho, M., Song, M., Comuzzi, M., Yoo, S.: Evaluating the effect of best
  practices for business process redesign: An evidence-based approach based on
  process mining techniques. Decision Support Systems  \textbf{104},  92--103
  (2017). \doi{10.1016/j.dss.2017.10.004}

\bibitem{https://doi.org/10.4121/uuid:3926db30-f712-4394-aebc-75976070e91f}
van Dongen, B.: Bpi challenge 2012 (2012).
  \doi{10.4121/uuid:3926db30-f712-4394-aebc-75976070e91f}

\bibitem{https://doi.org/10.4121/uuid:5f3067df-f10b-45da-b98b-86ae4c7a310b}
van Dongen, B.: Bpi challenge 2017 (2017).
  \doi{10.4121/uuid:5f3067df-f10b-45da-b98b-86ae4c7a310b}

\bibitem{10.1145/3576047}
Dumas, M., Fournier, F., Limonad, L., Marrella, A., Montali, M., Rehse, J.R.,
  Accorsi, R., Calvanese, D., De~Giacomo, G., Fahland, D., Gal, A., La~Rosa,
  M., V\"{o}lzer, H., Weber, I.: Ai-augmented business process management
  systems: A research manifesto. ACM Trans. Manage. Inf. Syst.  \textbf{14}(1)
  (Jan 2023). \doi{10.1145/3576047}

\bibitem{Dumas2018}
Dumas, M., La~Rosa, M., Mendling, J., Reijers, H.A.: Process Redesign, pp.
  297--339. Springer Berlin Heidelberg, Berlin, Heidelberg (2018).
  \doi{10.1007/978-3-662-56509-4_8}

\bibitem{edge2025localglobalgraphrag}
Edge, D., Trinh, H., Cheng, N., Bradley, J., Chao, A., Mody, A., Truitt, S.,
  Metropolitansky, D., Ness, R.O., Larson, J.: From local to global: A graph
  rag approach to query-focused summarization (2025),
  \url{https://arxiv.org/abs/2404.16130}

\bibitem{DBLP:journals/dss/FehrerFLRW22}
Fehrer, T., Fischer, D.A., Leemans, S.J.J., R{\"{o}}glinger, M., Wynn, M.T.: An
  assisted approach to business process redesign. Decis. Support Syst.
  \textbf{156},  113749 (2022). \doi{10.1016/J.DSS.2022.113749}

\bibitem{DBLP:conf/hicss/GrossMM19}
Gro{\ss}, S., Malinova, M., Mendling, J.: Navigating through the maze of
  business process change methods. In: 52nd Hawaii International Conference on
  System Sciences, {HICSS} 2019, Grand Wailea, Maui, Hawaii, USA, January 8-11,
  2019. pp. 1--10. ScholarSpace (2019),
  \url{https://hdl.handle.net/10125/60061}

\bibitem{Klievtsova_2026}
Klievtsova, N., Kampik, T., Mangler, J., Rinderle-Ma, S.: Conversational
  process model redesign. International Journal of Cooperative Information
  Systems  \textbf{35}(01) (Mar 2026). \doi{10.1142/s0218843026500048}

\bibitem{10.1007/978-3-031-26886-1_23}
L{\'o}pez-Pintado, O., Halenok, I., Dumas, M.: Prosimos: Discovering
  and simulating business processes with differentiated resources. In:
  Enterprise Design, Operations, and Computing. EDOC 2022 Workshops. pp.
  346--352. Springer International Publishing, Cham (2023)

\bibitem{MANSAR2005457}
Mansar, S.L., Reijers, H.A.: Best practices in business process redesign:
  validation of a redesign framework. Computers in Industry  \textbf{56}(5),
  457--471 (2005). \doi{10.1016/j.compind.2005.01.001}

\bibitem{Maruster2009}
Măruşter, L., van Beest, N.: Redesigning business processes: A methodology
  based on simulation and process mining techniques. Knowledge and Information
  Systems  \textbf{21},  267--297 (12 2009). \doi{10.1007/s10115-009-0224-0}

\bibitem{10.1007/978-3-642-21759-3_7}
Niedermann, F., Schwarz, H.: Deep business optimization: Making business
  process optimization theory work in practice. In: Enterprise,
  Business-Process and Information Systems Modeling. pp. 88--102. Springer
  Berlin Heidelberg, Berlin, Heidelberg (2011)

\bibitem{10387715}
Pan, S., Luo, L., Wang, Y., Chen, C., Wang, J., Wu, X.: Unifying large language
  models and knowledge graphs: A roadmap. IEEE Transactions on Knowledge and
  Data Engineering  \textbf{36}(7),  3580--3599 (2024).
  \doi{10.1109/TKDE.2024.3352100}

\bibitem{DBLP:journals/procsci/RebmannSGA25}
Rebmann, A., Schmidt, F.D., Glavas, G., van~der Aa, H.: On the potential of
  large language models to solve semantics-aware process mining tasks. Process
  Sci.  \textbf{2}(1) (2025). \doi{10.1007/S44311-025-00019-3}

\bibitem{REIJERS2005283}
Reijers, H., {Liman Mansar}, S.: Best practices in business process redesign:
  an overview and qualitative evaluation of successful redesign heuristics.
  Omega  \textbf{33}(4),  283--306 (2005). \doi{10.1016/j.omega.2004.04.012}

\bibitem{VANDERAALST2001451}
{van der Aalst}, W.: Re-engineering knock-out processes. Decision Support
  Systems  \textbf{30}(4),  451--468 (2001).
  \doi{10.1016/S0167-9236(00)00136-6}

\bibitem{VANDUN2023113880}
{van Dun}, C., Moder, L., Kratsch, W., Röglinger, M.: Processgan: Supporting
  the creation of business process improvement ideas through generative machine
  learning. Decision Support Systems  \textbf{165},  113880 (2023).
  \doi{10.1016/j.dss.2022.113880}

\bibitem{10.1007/978-3-540-72988-4_40}
Weber, B., Rinderle, S., Reichert, M.: Change patterns and change support
  features in process-aware information systems. In: Advanced Information
  Systems Engineering. pp. 574--588. Springer Berlin Heidelberg, Berlin,
  Heidelberg (2007)

\bibitem{10.5555/3600270.3602070}
Wei, J., Wang, X., Schuurmans, D., Bosma, M., Ichter, B., Xia, F., Chi, E.H.,
  Le, Q.V., Zhou, D.: Chain-of-thought prompting elicits reasoning in large
  language models. In: Proceedings of the 36th International Conference on
  Neural Information Processing Systems. NIPS '22, Curran Associates Inc., Red
  Hook, NY, USA (2022)

\end{thebibliography}

\end{document}